\documentclass[twoside]{article}
\usepackage{subfiles}
\usepackage[accepted]{aistats2019}
\usepackage{hyperref}
\usepackage[utf8]{inputenc}

\usepackage{csquotes}
\usepackage[style=authoryear,
            sorting=nyt,
            maxcitenames=2,
            maxbibnames=4,
            uniquelist=false,
            natbib=true, 
            url=false,
            doi=false,
            eprint=true,
            isbn=false,
            dashed=false]{biblatex}
\usepackage{amsfonts,amsmath,amssymb,amsthm}       

\usepackage{microtype,marvosym}

\usepackage{empheq}           %
\usepackage[most]{tcolorbox}  
\tcbuselibrary{theorems}      %

\usepackage{menukeys}

\usepackage{tikz,pgfplots}
\usepackage{subcaption}

\usepackage{cancel}     
\usepackage{sectsty}    
\usepackage{pifont}
\pgfplotsset{compat=newest}
\pgfplotsset{plot coordinates/math parser=false}

\usetikzlibrary{arrows,shapes,plotmarks,pgfplots.groupplots,pgfplots.colormaps,shadows,graphs,fadings}

\tikzset{>=stealth'} 
\tikzstyle{graphnode} = 
   [circle,draw=black,minimum size=22pt,text centered,text
     width=22pt,inner sep=0pt] 
\tikzstyle{var}   =[graphnode,fill=white]
\tikzstyle{obs}   =[graphnode,fill=black,text=white]
\tikzstyle{fac}   =[rectangle,draw=black,fill=black!25,minimum size=5pt]
\tikzstyle{facprior} =[rectangle,draw=black,fill=black,text=white,minimum size=5pt]
\tikzstyle{edge}  =[double=black,-]
\tikzstyle{prior} =[rectangle, draw=black, fill=black, minimum size=
5pt, inner sep=0pt]
\tikzstyle{dirprior} = [circle, draw=black, fill=black, minimum
size=5pt, inner sep=0pt]
\tikzstyle{dir}  =[double=black,->]
\pgfplotsset{compat=newest}
\pgfplotsset{
  every axis legend/.append style =
    {
      cells = { anchor = east },
      draw  = none
    },
}  
\pgfkeys{/pgfplots/mystyle/.style={
  semithick,
  tick style={major tick length=4pt,semithick,gray},
  xtick align = inside,
  ytick align = inside
  }}
  
\makeatletter
\newcommand{\uRightarrow}[2][]{\ext@arrow 0359\Rightarrowfill@{#1}{#2}}
\makeatother

\definecolor{lred}{RGB}{200,0,0}
\definecolor{dred}{RGB}{130,0,0} 
\definecolor{dblu}{RGB}{0,0,130}
\definecolor{dgre}{RGB}{0,130,0} 
\definecolor{dgra}{RGB}{50,50,50}
\definecolor{mgra}{RGB}{221,222,214}
\definecolor{lgra}{RGB}{238,238,234}
\definecolor{MPG}{RGB}{000,125,122}
\definecolor{lMPG}{RGB}{000,190,189}
\definecolor{ora}{HTML}{FF9933} 
\definecolor{lblu}{HTML}{7DA7D9}

\definecolor{TUred}{RGB}{141,45,57}
\definecolor{TUdark}{RGB}{51,51,51}
\definecolor{TUgold}{RGB}{174,159,109}
\definecolor{TUgray}{RGB}{175,179,183}

\newtcbox{\mybox}[1][]{%
    nobeforeafter, math upper, tcbox raise base,
    enhanced, colframe=blue!30!black,
    colback=ora!100, boxrule=1pt,
    #1}

\newtcbtheorem[number within=section]{boxtheorem}{Theorem}%
{colback=green!5,colframe=green!35!black,fonttitle=\bfseries}{th}

\newcommand{\grad}{\nabla}

\newcommand{\g}{\,|\,}

\newcommand{\Dset}{\mathcal{D}}
\newcommand{\Exp}{\mathbb{E}}

\newcommand{\cov}{\operatorname{cov}}

\renewcommand{\Re}{\mathbb{R}}

\newcommand{\diag}{\operatorname{diag}}
 
\newcommand{\N}{\mathcal{N}}

\newcommand{\B}{\mathcal{B}} 

\renewcommand{\L}{\mathcal{L}}
\newcommand{\Trans}{^{\intercal}}
  
\newcommand{\argmin}{\operatorname*{arg\:min}}

\renewcommand{\vec}{\boldsymbol} 
 
\newcommand{\vect}[1]{\overrightarrow{#1}}

\renewcommand{\O}{\mathcal{O}}

\newcommand{\Id}{\vec{I}}

\renewcommand{\L}{\mathcal{L}}

\usepackage{colonequals}
\newcommand{\ce}{\colonequals}

\DeclareSymbolFont{stmry}{U}{stmry}{m}{n}
\DeclareMathSymbol\leftarrowtriangle\mathrel{stmry}{"5E}
\DeclareMathSymbol\rightarrowtriangle\mathrel{stmry}{"5F}
\DeclareMathSymbol\leftrightarrowtriangle\mathrel{stmry}{"5D}
\DeclareMathSymbol\obar\mathrel{stmry}{"11}
\DeclareMathSymbol\otimes\mathrel{stmry}{"0F}
\DeclareMathSymbol\ominus\mathrel{stmry}{"17}
\DeclareMathSymbol\sslash\mathrel{stmry}{"0C}

\renewcommand{\to}{\operatorname*{\rightarrowtriangle}}

\makeatletter
\DeclareRobustCommand{\symkron}{%
  \mathbin{\mathpalette\o@minus@times\relax}%
}
\newcommand{\o@minus@times}[2]{%
  \ooalign{$\m@th#1\ominus$\cr$\m@th#1\otimes$\cr}%
}
\makeatother

\makeatletter
\DeclareRobustCommand{\asymkron}{%
  \mathbin{\mathpalette\o@bar@times\relax}%
}
\newcommand{\o@bar@times}[2]{%
  \ooalign{$\m@th#1\obar$\cr$\m@th#1\otimes$\cr}%
}
\makeatother

\usepackage{mathtools}
\usepackage{algorithm}
\usepackage{algpseudocode}
\algrenewcommand{\algorithmiccomment}[1]{\hfill {\footnotesize $\sslash$ #1}}
\algrenewcommand{\alglinenumber}[1]{\tt\scriptsize #1}

\makeatletter
\@addtoreset{algorithm}{chapter} 
\makeatother

\makeatletter
\def\therule{\makebox[\algorithmicindent][l]{\hspace*{.5em}\vrule height .75\baselineskip depth .25\baselineskip}}%

\newtoks\therules
\therules={}
\def\appendto#1#2{\expandafter#1\expandafter{\the#1#2}}
\def\gobblefirst#1{
  #1\expandafter\expandafter\expandafter{\expandafter\@gobble\the#1}}%
\def\LState{\State\unskip\the\therules}
\def\pushindent{\appendto\therules\therule}%
\def\popindent{\gobblefirst\therules}%
\def\printindent{\unskip\the\therules}%
\def\printandpush{\printindent\pushindent}%
\def\popandprint{\popindent\printindent}%

\algdef{SE}[WHILE]{While}{EndWhile}[1]
  {\printandpush\algorithmicwhile\ #1\ \algorithmicdo}
  {\popandprint\algorithmicend\ \algorithmicwhile}%
\algdef{SE}[FOR]{For}{EndFor}[1]
  {\printandpush\algorithmicfor\ #1\ \algorithmicdo}
  {\popandprint\algorithmicend\ \algorithmicfor}%
\algdef{S}[FOR]{ForAll}[1]
  {\printindent\algorithmicforall\ #1\ \algorithmicdo}%
\algdef{SE}[LOOP]{Loop}{EndLoop}
  {\printandpush\algorithmicloop}
  {\popandprint\algorithmicend\ \algorithmicloop}%
\algdef{SE}[REPEAT]{Repeat}{Until}
  {\printandpush\algorithmicrepeat}[1]
  {\popandprint\algorithmicuntil\ #1}%
\algdef{SE}[IF]{If}{EndIf}[1]
  {\printandpush\algorithmicif\ #1\ \algorithmicthen}
  {\popandprint\algorithmicend\ \algorithmicif}%
\algdef{C}[IF]{IF}{ElsIf}[1]
  {\popandprint\pushindent\algorithmicelse\ \algorithmicif\ #1\ \algorithmicthen}%
\algdef{Ce}[ELSE]{IF}{Else}{EndIf}
  {\popandprint\pushindent\algorithmicelse}%
\algdef{SE}[PROCEDURE]{Procedure}{EndProcedure}[2]
   {\printandpush\algorithmicprocedure\ \textproc{#1}\ifthenelse{\equal{#2}{}}{}{(#2)}}%
   {\popandprint\algorithmicend\ \algorithmicprocedure}%
\algdef{SE}[FUNCTION]{Function}{EndFunction}[2]
   {\printandpush\algorithmicfunction\ \textproc{#1}\ifthenelse{\equal{#2}{}}{}{(#2)}}%
   {\popandprint\algorithmicend\ \algorithmicfunction}%
\makeatother

\begin{document}

%

%




\twocolumn[

\runningtitle{Active Probabilistic Inference on Matrices for Pre-Conditioning in Stochastic Optimization}
\aistatstitle{Active Probabilistic Inference on Matrices\\ for Pre-Conditioning in Stochastic Optimization}

\aistatsauthor{Filip de Roos \And Philipp Hennig}
\aistatsaddress{Max Planck-Institute for Intelligent Systems \and University of T\"ubingen, Germany\\
\texttt{[filip.de.roos|ph]@tue.mpg.de}}]


\begin{abstract}
  Pre-conditioning is a well-known concept that can significantly improve the convergence of optimization algorithms. For noise-free problems, where good pre-conditioners are not known a priori, iterative linear algebra methods offer one way to efficiently construct them. For the \emph{stochastic} optimization problems that dominate contemporary machine learning, however, this approach is not readily available. We propose an iterative algorithm inspired by classic iterative linear solvers that uses a probabilistic model to \emph{actively} infer a pre-conditioner in situations where Hessian-projections can only be constructed with strong Gaussian noise. The algorithm is empirically demonstrated to efficiently construct effective pre-conditioners for stochastic gradient descent and its variants. Experiments on problems of comparably low dimensionality show improved convergence. In very high-dimensional problems, such as those encountered in deep learning, the pre-conditioner effectively becomes an automatic learning-rate adaptation scheme, which we also empirically show to work well.
\end{abstract}

\section{INTRODUCTION} 
\label{sec:introduction}

Contemporary machine learning heavily features empirical risk minimization, on loss functions of the form
\begin{equation}\label{eq:full-risk}
  \L(w) = \frac{1}{|\Dset|} \sum_{i\in \Dset} ^{|\Dset|} l_i(w),
\end{equation}
where $\Dset$ is a data-set and $w$ are parameters of the optimization problem (e.g.~the weights of a neural network). Data sub-sampling in such problems gives rise to stochastic optimization problems: If gradients are computed over a randomly sampled batch $\B\subset \Dset$ of data-points, they give rise to a stochastic gradient
\begin{equation}
  \grad \tilde{\mathcal{L}} (w)= \frac{1}{|\B |}\sum\limits_{i\in \B}^{|\B|} \grad l_i(w).
  \label{eq:batch-risk}
\end{equation}
If $\Dset$ is large and $\B$ is sampled iid., then $\grad\tilde{\L}$ is an unbiased estimator and, by the multivariate Central Limit Theorem, is approximately Gaussian distributed (assuming $|\B|\ll|\Dset|$) around the full-data gradient as \citep[e.g.][\textsection 2.18]{balles2017coupling,vaart_1998}
\begin{equation}\label{eq:grad_likelihood}
  p\left(\grad \tilde{\mathcal{L}} (w)\big{|} \grad \L\right) = \N\left(\grad\tilde{\L}(w); \grad\L(w), |\B|^{-1}  \cov(\grad \L)\right)
\end{equation}
where $\cov(\nabla\L)$ is the empirical covariance over $\Dset$. For problems of large scale in both data and parameter-space, stochastic optimization using stochastic gradient descent \citep{robbins1951stochastic} and its by now many variants (e.g., momentum \citep{polyak1964some}, Adam \citep{kingma2014adam}, etc.) are standards.

For (non-stochastic) gradient descent, it is a classic result that convergence depends on the condition-number of the objective's Hessian $B(w)\ce\nabla\nabla\Trans \L(w)$. For example, Thm.~3.4 in Nocedal \& Wright \citep{NoceWrig06} states that the iterates of gradient descent with optimal local step sizes on a twice-differentiable objective $\L(w):\Re^N\to\Re$ converge to a local optimum $w_*$ such that, for sufficiently large $i$,
\begin{multline}
  \L(w_{i+1}) - \L(w_*) \leq r^2 (\L(w_{i}) - \L(w_*)), \quad \\ \text{with}\quad r\in \left(\frac{\lambda_N - \lambda_1}{\lambda_N + \lambda_1},1\right),
\end{multline}
where $\lambda_N$ and $\lambda_1$ are the largest and smallest eigen-value of the Hessian $B(w)$, respectively. In noise-free optimization, it is thus common practice to try and reduce the condition number $\kappa\ce\lambda_N/\lambda_1$ of the Hessian, by a linear re-scaling $\bar{w}=P\Trans w$ of the input space using a \emph{pre-conditioner} $P\in \Re^{N\times N}$.

For ill-conditioned problems ($\kappa \gg 1$), effective pre-conditioning can drastically improve the convergence. Choosing a good pre-conditioner is a problem-dependent art. A good pre-conditioner should decrease the condition number of the problem and be cheap to apply, either through sparseness or low-rank structure.

Sometimes pre-conditioners can be constructed as a good `a-priori' guess of the inverse Hessian $B^{-1}$. I no such good guess is available, then 
in the deterministic/non-stochastic setting, iterative linear-algebra methods can be used to build a low-rank pre-conditioner. For example, if $P$ spans the leading eigen-vectors of the Hessian, the corresponding eigen-values can be re-scaled to change the spectrum, and by extension the condition number. Iterative methods such as those based on the Lanczos process \citep[][\textsection 10.1]{golub2012matrix} try to consecutively expand a helpful subspace by choosing $B$-conjugate vectors, which can fail in the presence of inexact computations. Due to these intricate instabilities \citep[][p.~282]{trefethen1997numerical} such algorithms tend not to work with the level of stochasticity encountered in practical machine learning applications. 

Below, we propose a framework for the efficient construction of pre-conditioners in settings with noise-corrupted Hessian-vector products available.
Our algorithm consists of three main components: We first build a probabilistic Gaussian inference model for matrices from noisy matrix-vector products (Section~\ref{sec:matrix_inference}) by extending existing work on matrix-variate Gaussian inference. Then we construct an active algorithm that selects informative vectors aiming to explore the Hessian's dominant eigen-directions (Section~\ref{sec:algorithm}). The structure of this algorithm is inspired by that of the classic Arnoldi and Lanczos iterations designed for noise-free problems. Finally (Section~\ref{sub:algorithmic_details}), we provide some ``plumbing'' to empirically estimate hyper-parameters and efficiently construct a low-rank pre-conditioner and extend the algorithm to the case of high-dimensional models (Section~\ref{sub:high_dimensional_modification}). We evaluate the algorithm on some simple experiments to empirically study its properties as a way to construct pre-conditioners and test it on both a low-dimensional\footnote{\href{https://github.com/fderoos/probabilistic_hessian}{Code repository.}}, and a high-dimensional deep learning problem 
. While we use pre-conditioning as the principal setting, the main contribution of our framework is the ability to construct matrix-valued estimates in the presence of noise, and to do so at complexity \emph{linear} in the width and height of the matrix. It is thus applicable to problems of large scale, also in domains other than pre-conditioning and optimization in general. In contrast to a simple averaging of random observations, our algorithm actively chooses projections in an effort to improve the estimate.
\subsection{Related Work} 
\label{sec:background_related_work}

Our approach is related to optimization methods that try to emulate the behaviour of Newton's method without incurring its cubic per-step cost. That is, iterative optimization updates
$w_{i+1} = w_i - d_i $ that try to find a search direction $d_i$ that is an approximate solution to the linear problem
\begin{equation}
  B(w_i) d_i = \nabla \L(w_i).
  \label{eq:newton}
\end{equation}
This includes quasi-Newton methods like BFGS and its siblings \citep{dennis1977quasi}, and Hessian-free optimization \citep{pearlmutter1994fast,Martens2010}. These methods try to keep track of the Hessian during the optimization. Pre-conditioning is a simpler approach that separates the estimation of the (inverse) Hessian from the on-line phase of optimization and moves it to an initialization phase. Our algorithm could in principle be run in every single step of the optimizer, such as in Hessian-free optimization. However, this would multiply the incurred cost, which is why we here only study its use for pre-conditioning.

There are stochastic variants of quasi-Newton methods and other algorithms originally constructed for noise-free optimization \citep[e.g.][]{schraudolph2007stochastic,byrd2016stochastic}. These are generally based on collecting independent random (not actively designed) samples of quasi-Newton updates. Estimates can also be obtained by regularizing estimates \citep[e.g.][]{2018arXiv180204310W} or partly update the stochastic gradient by reusing elements of a batch \citep[e.g.][]{2018arXiv180205374B}. The conceptual difference between these methods and ours is that we actively try to design informative observations by explicitly taking evaluation uncertainty into account.

Our inference scheme is an extension of Gaussian models for inference on matrix elements, which started with early work by \citet{DAWID01041981} and was recently extended in the context of probabilistic numerical methods \citep{hennig2013quasi,hennig2015,2018arXiv180204310W}. Our primary addition to these works is the algebra required for dealing with structured observation noise.

\section{THEORY} 
\label{sec:theory}

Our goal in this work is to construct an active inference algorithm for low-rank pre-conditioning matrices that can deal with data in the form of Hessian-vector multiplications corrupted by significant (not just infinitesimal) Gaussian noise. To this end, we will adopt a probabilistic viewpoint with a Gaussian observation likelihood, and design an active evaluation policy that aims to efficiently collect informative, non-redundant Hessian-vector products. The algorithm will be designed so that it produces a Gaussian posterior measure over the Hessian of the objective function, such that the posterior mean is a low-rank matrix.

\subsection{Matrix Inference} 
\label{sec:matrix_inference}

Bayesian inference on matrices $B\in \Re ^{N\times N}$ can be realised efficiently in a Gaussian framework by re-arranging the matrix elements into a vector $\vect{B}\in \Re^{N^2\times 1}$, then performing standard Gaussian inference on this vector \citep[e.g.][]{DAWID01041981}.
Although Hessian matrices are square and symmetric, the following derivations apply equally well to rectangular matrices. There are specializations for symmetric matrices~\citep{hennig2015}, but since they significantly complicate the derivations below, we use this weaker model.

Assume that we have access to observations $Y\in\Re^{N\times m}$ of matrix-vector products $Y=BS$ along the search directions $S\in\Re^{N\times m}$. In the vectorized notation, this amounts to a linear projection of $\vect{B}$ through the Kronecker product matrix $(I\otimes S\Trans)$:
\begin{equation*}
  Y_{ab}= (I\otimes S\Trans)\vect{B} =\sum_{ij} \delta_{ai}B_{ij}S_{jb}=[BS]_{ab}.
\end{equation*}
If the observations are exact (noise-free), the likelihood function is a Dirac distribution,
\begin{equation}\label{eq:delta-likelihood}
  \begin{split}
    p(Y\g B,S)&=\delta(\vect{Y}-(I\otimes S\Trans)B)=\\ &=\lim_{\beta \rightarrow 0} \N(Y; (I\otimes S\Trans)B, \beta \Lambda_0).
  \end{split}
\end{equation}
For conjugate inference, we assign a Gaussian prior over $B$, with a prior mean matrix $B_0$ and a covariance consisting of a Kronecker product of 2 symmetric positive-definite matrices.
\begin{multline}
  \N(B,B_0,V\otimes W)=\frac{1}{((2\pi)^{n^2}/|V|^n|W|^n)^{1/2}} \\ \cdot \exp \left(-\frac{1}{2} (\vect{B}-\vect{B_0})\Trans (V\otimes W)^{-1} (\vect{B}-\vect{B_0}) \right).
  \label{eq:general_prior}
\end{multline}
This is an equivalent re-formulation of the \emph{matrix-variate Gaussian} distribution~\citep{DAWID01041981}. For simplicity, and since we are inferring a Hessian matrix $B$ (which is square and symmetric), we set $V=W$. This combination of prior \eqref{eq:general_prior} and likelihood \eqref{eq:delta-likelihood} has previously been discussed \citep[detailed derivation, e.g., in][]{hennig2015}. It gives rise to a Gaussian posterior distribution, whose mean matrix is given by
\begin{equation}
  \begin{split}
    B_m= &B_0 + (W\otimes W)(I\otimes S)\vect{X}= B_0+WX S\Trans W =\\ &B_0 + (Y-B_0 S)(S\Trans W S)^{-1} S\Trans W
    \label{eq:posterior_mean}
  \end{split}
\end{equation}
where $X$ is found as the solution to the linear system
\begin{equation}
  (\underbrace{W\otimes S\Trans WS}_{\equalscolon G})\vect{X}=\vect{\underbrace{(Y-B_0 S)}_{\equalscolon\vect{\Delta}}},
  \label{eq:X_noise_free}
\end{equation}
using the Kronecker product's property that $(A\otimes B)^{-1}=A^{-1}\otimes B^{-1}$ (note that the matrix $(S\Trans W S)\in\Re^{m\times m}$ can be inverted in $\mathcal{O}(m^3)$).


\subsubsection{Adding Noise} 
\label{sec:adding_noise}

While noise-free observations of Gaussian-distributed matrices have been studied before, observation noise beyond the fully iid.~case, \citep[cf.][]{2018arXiv180204310W} is a challenging complication. A lightweight inference-scheme for noisy observations is one of the core contributions of this paper.
In empirical risk minimization problems, mini-batching replaces a large or infinite sum (the population risk, Eq.~\eqref{eq:full-risk}) into a small sum of individual loss functions (Eq.~\eqref{eq:batch-risk}). Analogous to Eq.~\eqref{eq:grad_likelihood}, the Hessian $\tilde{B}$ of the batch risk $\tilde\L$ is thus corrupted relative to the true Hessian by $B$ a Gaussian likelihood:
\begin{multline}
  \tilde{B}(w) = B(w) + \Gamma \quad \text{with} \quad \Gamma \sim \N(0,\Sigma)\quad \\ \text{and}\quad \Sigma = \cov(\nabla\nabla\Trans \L) / |\B|.
\end{multline}
If we now compute matrix-vector products of this batch Hessian with vectors $S\in\Re^{N\times m}$, even under the simplifying assumption of Kronecker structure $\Sigma=\Lambda\otimes\Lambda$ in the covariance, the observation likelihood becomes

\begin{equation}\label{eq:likelihood-noise}
   p(Y\g B ,S)=\N(\vect{Y}; \vect{BS}, \underbrace{\Lambda\otimes (S\Trans\Lambda S)_{ii}}_{\equalscolon R}).
 \end{equation}
The subscript $ii$ in Eq.~\eqref{eq:likelihood-noise} is there to highlight the diagonal structure of the right matrix due to the independent batches we use to calculate the Hessian-vector products. Relative to Eq~\eqref{eq:X_noise_free}, this changes the Gram matrix to be inverted from $G$ to $(G + R)$.

To get the posterior of $B$ in the noisy setting after $m$ observations, instead of Eq.~\eqref{eq:X_noise_free}, we now have to solve the linear problem
\begin{equation}\label{eq:X_noise}
  \underbrace{(G+R)}_{\Re^{nm\times nm}}\vect{X}=\vect{\Delta}.
\end{equation}
This is a more challenging computation, since the sum of Kronecker products does not generally have an analytic inverse. However, Eq.~\eqref{eq:X_noise} is a so-called matrix pencil problem, which can be efficiently addressed using a generalized eigen-decomposition \citep[e.g.][\textsection 7.7]{golub2012matrix}. That is, by a matrix $U$ and diagonal matrix $D=\diag(\vec{\lambda})$ such that

\begin{equation}
   GV=DR V \qquad\text{with}\qquad V\Trans R V=I.
   \label{eq:GEV}
 \end{equation}
Eigen-decompositions distribute over a Kronecker product \citep{loan00kronec}. And this property is inherited by the generalized eigen-decomposition, which offers a convenient way to rewrite a matrix in terms of the other. Eq.~\eqref{eq:X_noise} can thus be solved with two generalized eigen-decompositions. The left and right parts of the Kronecker products of Eq.~\eqref{eq:X_noise} are written with separate generalized eigen-decompositions as
\begin{xalignat*}{2}
  WU=& \Lambda UD, & U\Trans \Lambda U &= I\\
  (S\Trans WS) V=& (S\Trans \Lambda S)_{ii} V\Omega, &V\Trans (S\Trans \Lambda S)_{ii} V&=I .
  \label{eq:generalised_eigen-value}
\end{xalignat*}
$U,D$ contain the generalized eigen-vectors and eigen-values from the left Kronecker term and $V,\Omega$ are analogous for the right Kronecker term.
\begin{align*}
  &\vect{\Delta}=(W\otimes S\Trans W S + \Lambda \otimes  (S\Trans \Lambda S)_{ii})\vect{X}\\
  &=(\Lambda U\otimes (S\Trans \Lambda S)_{ii}V) (D\otimes\Omega + I\otimes I)(U^{-1}\otimes V^{-1})\vect{X}\\
  &=(U^{-T}\otimes V^{-T} ) (D\otimes\Omega + I\otimes I)(U^{-1}\otimes V^{-1})\vect{X}
\end{align*}
In the first step above, the left matrix is expressed in terms of the right matrix in both terms by means of the generalized eigen-decomposition. In remaining step, the conjugacy property in Eq.~\eqref{eq:GEV} is used to simplify the expression to the inversion of a diagonal matrix and the Kronecker product of the generalized eigen-vectors.
The solution now becomes
\begin{equation}
  \begin{split}
      \vect{X} &=(U\otimes V)\underbrace{(D\otimes\Omega + I\otimes I)^{-1}\vect{U\Trans \Delta V}}_{\frac{1}{(D_{jj}\Omega_{ii}+1)}\odot U\Trans \Delta V = \Psi_{ji}} \\ &=(U\otimes V)\vect{\Psi} = U\Psi V\Trans
  \end{split}
  \label{eq:X_noise_solution}
\end{equation}
where $\odot$ refers to the Hadamard product of the two matrices. Using this form, we can represent the posterior mean estimate for $B$ with Eq.~\eqref{eq:posterior_mean} where $X$ is replaced with the solution from Eq.~\eqref{eq:X_noise_solution}.
\begin{equation*}
  B_m=B_0+\underbrace{WX}_{\Re^{N\times m}} \underbrace{S\Trans W}_{\Re^{m\times N}}
\end{equation*}
This matrix can generally not be written in a simpler analytic form. If the prior mean $B_0$ is chosen as a simple matrix (e.g.~a scaled identity), then $B_m$ would admit fast $\mathcal{O}(Nm)$ multiplication and inversion (using the matrix inversion lemma) due to its low-rank outer-product structure.


\subsection{Active Inference} 
\label{sec:algorithm}

The preceding section constructed an inference algorithm that turns noisy projections of a matrix into a low-rank estimate for the latent matrix. The second ingredient of our proposed algorithm, outlined in this section, is an active policy that chooses non-redundant projection directions to efficiently improve the posterior mean. Algorithm~\ref{alg} summarizes as pseudo-code.

The structure of this algorithm is motivated, albeit not exactly matched to, that of stationary iterative linear solvers and eigen-solvers such as GMRES \citep{saad1986gmres} and Conjugate gradients \citep{hestenes1952methods} (and the corresponding eigen-value-iterations, the Arnoldi process and the Lanczos process. Cf.~\citep[][\textsection 10]{golub2012matrix} and \citep[][\textsection VI]{trefethen1997numerical}). These algorithms can be interpreted as optimization methods that iteratively expand a low-rank approximation to (e.g.~in the case of Conjugate gradients / Lanczos) the Hessian of a quadratic problem, then solve the quadratic problem within the span of this approximation. In our algorithm, the exact low-rank approximation is replaced by the posterior mean \emph{estimate} arising from the Bayesian inference routine described in Section~\ref{sec:matrix_inference}. This leads the algorithm to suppress search directions that are co-linear with those collected in previous iterations, focussing instead on the efficient collection of new information.

Readers familiar with linear solvers like conjugate gradients will be able to recognise the structural similarity of Algorithm~\ref{alg} to linear solvers, with two differing aspects. Each iteration constructs a projection direction $s_i$, collects one matrix-vector multiplication, $y_i=\tilde{B}s_i$ and rescales them by a step size $\beta_i$ (here set to 1 and omitted). A linear solver would update the solution $x_i$ and residual $r_i$ using $s_i$, $y_i$ and $\beta_i$ but we let the algorithm stay at $x_0$ and sample new search directions and projections. The core difference to a solver is in line~\ref{alg:important-line}: Where the classic solvers would perform a Gram-Schmidt step, we instead explicitly perform Gaussian inference on the Hessian $B$. In the noise-free limit the proposed method would choose the same search directions as projection methods, a superclass of iterative solvers containing algorithms such as GMRES and CG \citep{hennig2015}.

\subsection{Algorithmic Details}
\label{sub:algorithmic_details}

Like all numerical methods, our algorithm requires the tuning of some internal parameters and some implementational engineering. 
We set all free parameters in an empirical fashion, via cheap pre-computations, and use standard linear-algebra tools for the internals.

\subsubsection{Estimating Parameters} 
\label{sec:estimating_parameters}

The Gaussian prior~\eqref{eq:general_prior} and likelihood~\eqref{eq:likelihood-noise} (also used as parameters of Alg.~\ref{alg}) have several parameters. For simplicity and to limit computational cost, we set all these to scaled identity matrices: prior mean $B_0=b_0 I$, variance $W=w_0 I$ and noise covariance $\Lambda=\lambda_0 I$. We set these parameters empirically: Before the method tries to estimate the Hessian, it gathers gradients and Hessian-gradient products locally, from a small number of initial batches. Then we set the parameters of the prior over $B$ as the empirical estimates
\begin{equation}
   1/b_0 = \sqrt{\frac{s\Trans B s}{s\Trans BB s}} \qquad\text{and}\qquad w_0 = \frac{s\Trans B s}{s\Trans s}.
   \label{eq:parameters}
\end{equation}
The noise variance for the likelihood set by an empirical esimate
\begin{equation*}
  \lambda_0 =(\Exp [g^2] - (\underbrace{\Exp [g]}_{\bar{g}})^2 )/ \sqrt{ s\Trans s }.
\end{equation*}
Since batch elements can vary quite extremely, however, we use a more robust choice, by setting it to the median of the variance estimates.
The mean gradient ($\bar{g}$) from the initial sampling is used for the first iteration of the Hessian-inference scheme, line \ref{alg:initial_grad}. A new search direction along which a Hessian-vector product is to be computed, is obtained by applying the inverse of the current estimated Hessian to a stochastic gradient:
\begin{equation}
  s_{i+1}=-\hat{B}^{-1}_i \grad \tilde{\L}(w).
  \label{eq:search_direction}
\end{equation}
The estimated Hessian is updated by first solving Eq.~\eqref{eq:X_noise} and using the result in Eq.~\eqref{eq:posterior_mean} to get the posterior mean: It is a sum of the prior mean and a low-rank outer product of two $N\times m$ matrices, with $N$ the number of parameters and $m$ the number of observations/iterations. A diagonal prior mean offers efficient inversion of the estimated Hessian by the matrix inversion lemma, and subsequent multiplication in Eq.~\eqref{eq:search_direction}. 

In the experiments below, the algorithm is used to construct a pre-conditioner for stochastic optimizers. In this application, it runs at the beginning of an optimization process, collecting several batches ``in-place'', each time computing a noisy matrix-Hessian product. To find the dominant $k$ eigen-directions, especially if noise is significant, the solver usually requires a number $m>k$ of iterations, producing a posterior estimate $B_m$ of rank $m$. To reduce the computational cost and memory requirements in the subsequent actual optimization run, we reduce the rank of this approximation down to $k$ using a standard fast singular value decomposition and obtain the singular values $\Sigma$ and the left singular vectors $U$ and estimate the Hessian $B\approx U\Sigma U\Trans$. This is equivalent to taking the symmetric part of a polar decomposition which yields the closest symmetric approximation to a matrix in Frobenius norm \citep{HIGHAM1988103}.
For an $N$-dimensional problem and a pre-conditioner of rank $k$, this method requires the storage of $\O (Nk)$ numbers and has a computational overhead of $\O (Nk)$ additional FLOPS compared to a standard sgd update. Such overhead is negligible for batch-sizes $|\B|> k$, the typical case in machine learning.

\begin{algorithm}[H]
\caption{Active probabilistic solver for linear problems of the form $Bx=b$, where multiplications with the matrix $B$ can only be performed corrupted by noise. When used in optimization, the target vector $b$ is set to the (noisy) gradient $\nabla\tilde\L$ (which amounts to searching for the Newton direction).}
\begin{algorithmic}[1]
\Procedure{InfHess}{$x_0$,$B(\cdot)$, $b$, $p(B)$, $p(Y\g S,B)$}
\LState $\phantom{H_{i}}\mathllap{r_0} = B_\text{true} \cdot x_0 - b$ \Comment{initial noisy gradient} \label{alg:initial_grad}
\For{$i = 1,\dots$}
\LState \raisebox{0pt}[-1pt][-1pt]{$\phantom{H_{i}}\mathllap{s_{i}} = -B_{i-1} ^{-1} r_{i-1}$} \Comment{step direction}
\LState $\phantom{H_{i}}\mathllap{y_{i}} = B_\text{true} \cdot s_{i}$ \Comment{\textbf{observe}}
\LState $\phantom{H_{i}}\mathllap{r_{i}} = B_\text{true} \cdot x_0 - b$ \Comment{new noisy gradient at $x_0$}
\LState $\phantom{H_{i}}\mathllap{B_{i}} = $ \Call{Infer}{$B\g Y_{i},S_{i},p(B), p(Y\g S,B)$}\label{alg:important-line} \setcounter{ALG@line}{6}
\LState \Comment{estimate $B$, using results from Section~\ref{sec:matrix_inference}}
\EndFor
\EndProcedure
\end{algorithmic}
\label{alg}
\end{algorithm}

\begin{figure*}[t]
  \centering \scriptsize
  \includegraphics{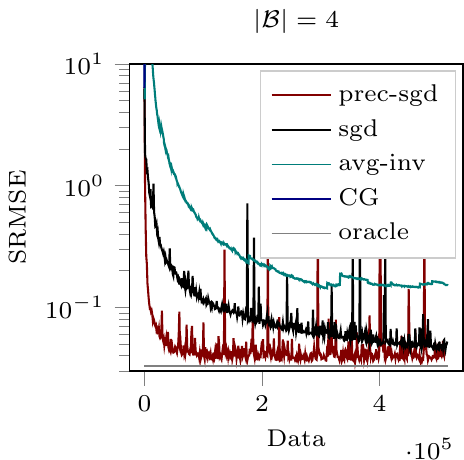}\hfill
  \includegraphics{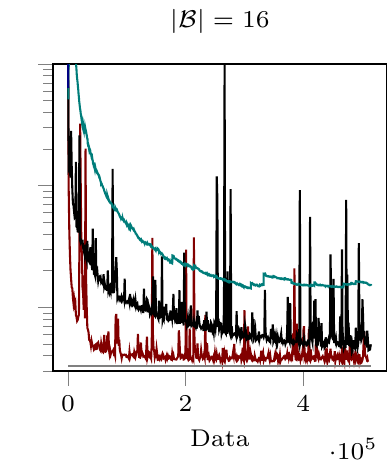}\hfill
  \includegraphics{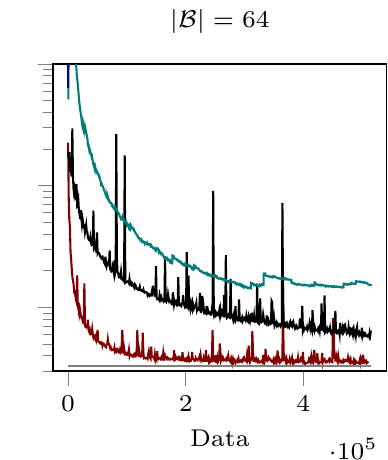}\hfill
  \includegraphics{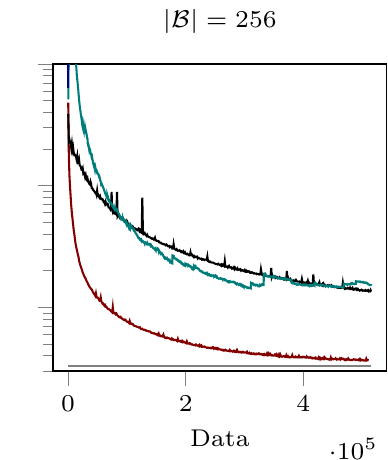}\hfill
  \caption{Comparison of sgd and pre-conditioned sgd on the linear test problem, along with other baselines. Details in text. The plots show data for four different choices of batch size $|\B|$ (and thus varying observation noise). For fairness to sgd, the abscissa is scaled in the number of data points loaded from disk (as opposed to the number of steps). Due to the noisy setting, vanilla CG is unstable and diverges in the first few steps.}
  \label{fig:ex_linear}
\end{figure*}

\begin{figure}[t]
  \centering \scriptsize
  \includegraphics{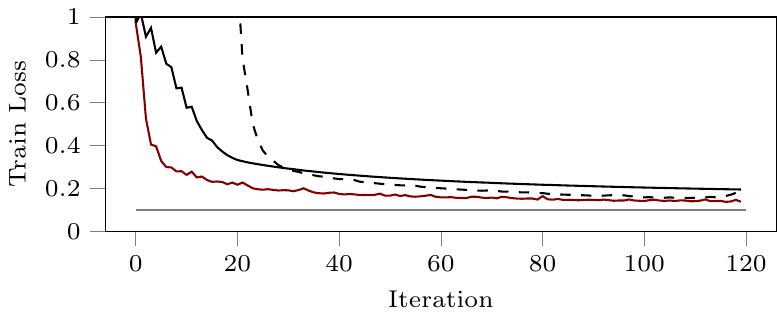}\\
  \includegraphics{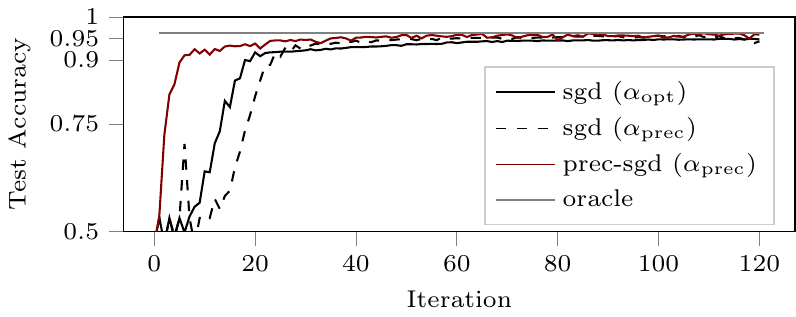}
  \caption{Progress of pre-conditioned and standard sgd on the MNIST logistic regression problem (details in text). Top plot shows progress on the raw optimization objective (train loss), the bottom plot shows generalization in terms of test accuracy. Two lines are shown for sgd. The solid one uses a step-size $\alpha_\text{opt}$ optimized for good performance. This directly induces a step-size $\alpha_\text{prec.}$ for the pre-conditioner. For comparison, we also show progress of sgd when directly using this step size---it makes sgd unstable.}
  \label{fig:ex_logistic}
\end{figure}

\subsubsection{Pre-conditioning} 
\label{sec:pre-conditioning}

A pre-conditioner $P$ is constructed to rescale the stochastic gradients in the direction of the singular vectors.
\begin{equation}
   P= \alpha (I + U[\beta I_k/ \sqrt{\Sigma} - I_k]U\Trans )
   \label{eq:P}
 \end{equation}
By rescaling the gradients with $P^2$, the linear system in Eq.~\eqref{eq:newton} is transformed into $P\Trans \tilde{B}(w_i)P P^{-1} d_i=P\Trans \grad \tilde{\L} (w_i)$. The goal, as outlined above, is to reduce the condition number of the transformed Hessian $P\Trans B(w_i) P$.
If the estimated vectors $U,\Sigma$ are indeed the real eigen-vectors and eigen-values, this approach would rescale these directions to have length $\beta$. Theoretically $\beta=1$ would be ideal if the real eigen-pairs are used. When instead an approximation of the subspace is used with poor approximations of the eigen-values $\tilde{\lambda}_i$, it is possible to scale a direction too much so the eigen-vectors corresponding to the largest eigen-values become the new smallest eigen-values. In our experiments this rarely happened because the Hessian contained many eigen-values $\lambda_i\ll 1$ and so $\beta=1$ could be used. 
The improved condition number allows usage of a larger step-size. We scale up by the empirical improvement of the condition number, i.e.~the fixed step-size of sgd $\eta$ is multiplied with $\alpha^2 =\Sigma_1/\Sigma_k$ (in Eq.~\eqref{eq:P}), the ratio between largest and the smallest estimated eigen-value. Using the current notation we can write the pre-conditioned sgd update as

\begin{equation*}
  w_{i+1}=w_i-\eta P^2 \nabla \tilde \L(w_i).
\end{equation*}

\subsection{High-Dimensional Modification} 
\label{sub:high_dimensional_modification}
Deep learning has become the benchmark for stochastic optimizers and it imposes many new constraints on the optimizers. Due to the large number of parameters, even if we would have access to the fraction of eigen-directions which make the Hessian ill-conditioned it would likely be inefficient to use because each vector has the same size as the network, and the approach would only work if a few of the eigen-values of the Hessian have a clear separation in magnitude. This changes the functionality we can expect from a pre-conditioner to keeping the parameter updates relevant with respect to the current condition number rather than finding all directions with problematic curvature. 

Some simplifications are required in order to adapt the algorithm for deep learning. The most important change is that we approximate the Hessian as a layer-wise block matrix, effectively treating each layer of a deep net as an independent task. Hessian-vector products are calculated using automatic differentiation \citep{pearlmutter1994fast}. Good estimates of the eigen-values for the algorithm proved difficult to get because of large deviations, likely due to the noisy Hessian-vector products. To alleviate this problem we changed the multiplicative update of the step-size presented in section \ref{sec:pre-conditioning} to redefining the step-length. Each time the algorithm is used to build a pre-conditioner, a new step-length is set to the scalar prior mean in Eq.~\eqref{eq:parameters}. The last modification is how the empirical parameters of the prior (section \ref{sec:estimating_parameters}) are set. $1/b_0$ is used as the new step-length and gave better results when a smaller step-size of $s\Trans B s/s\Trans B B s $ was used and $\lambda_0$ is estimated to $\lambda_0 =\sqrt{(\sum [g\Trans g] - \hat{g}\Trans \hat{g})/n}$, with $\hat{g}=\sum g_{(\B)}$. 
All the parameters of the prior and likelihood ($b_0$, $w_0$ and $\lambda_0$) are shared among the layers. No clear improvement was visible when treating the parameters separately for each layer.

\section{RESULTS} 
\label{sec:results}
We analyze and compare the performance of the proposed algorithm on a simple test problem along with two standard applications in machine learning.

\subsection{Regression} 
\label{sub:regression}

Figure~\ref{fig:ex_linear} shows results from a conceptual test setup designed to showcase the algorithm's potential: An ill-conditioned linear problem of a scale chosen such that the analytical solution can still be found for comparison. We used linear parametric least-squares regression on the SARCOS data-set \citep{VijayakumarS00} as a test setup. The data-set contains $|\Dset| = 44,484$ observations of a uni-variate\footnote{The data-set contains 7 such univariate target variables. Following convention, we used the first one.} response function $y_i=f(x_i)$ in a 21-dimensional space $x_i\in\Re^{21}$. We used the polynomial feature functions $\phi(x)=A [x,\operatorname{vec}(xx\Trans)] \in \Re^{253}$, with a linear mapping, $A$, manually designed to make the problem ill-conditioned. The model $f(x)=\phi(x)\Trans w$ with a quadratic loss function then yields a quadratic optimization problem,
\begin{multline}\label{eq:linearloss}
  w_* = \argmin_w \alpha\|w\|^2 +\frac{1}{|\Dset|} \| \Phi\Trans w - \vec{y} \|^2 =\\ \alpha\|w\|^2 + \frac{1}{|\Dset|} \sum_{i=1} ^{|\Dset|} (\phi(x_i)\Trans w - y_i)^2,
\end{multline}
where $\Phi\in\Re^{253 \times 44,484}$ is the map from weights to data. The exact solution of this problem is given by the regularized least-squares estimate $w_* = (\Phi\Phi\Trans/|\Dset| + \alpha\Id)^{-1} \Phi \vec{y} /|\Dset|$. For this problem size, this solution can be computed easily, providing an oracle baseline for our experiments. But if the number of features were higher (e.g.~$\gtrsim 10^4$), then exact solutions would not be tractable. One could instead compute, as in deep learning, batch gradients from Eq.~\eqref{eq:linearloss}, and also produce a noisy approximation of the Hessian $B=(\Phi\Phi\Trans /|\Dset| + \alpha\Id)$ as the low rank matrix
\begin{equation}\label{eq:batch-Hessian}
  \tilde{B} = \alpha\Id + \frac{1}{|\B|}\sum_{b\in\B} \phi(x_b)\phi(x_b)\Trans
\end{equation}
where $\B$ is a batch. Clearly, multiplying an arbitrary vector with this low-rank matrix has cost $\mathcal{O}(|\B|)$, thus providing the functionality required for our noisy solver. Figure~\ref{fig:ex_linear} compares the progress of vanilla sgd with that of pre-conditioned sgd if the pre-conditioner is constructed with our algorithm. In each case, the construction of the pre-conditioner was performed in 16 iterations of the inference algorithm. Even in the largest case of $\B=256$, this amounts to $4096$ data read, and thus only a minuscule fraction of the overall runtime.

An alternative approach would be to compute the inverse of $\tilde{B}$ separately for each batch, then average over the batch-least-squares estimate $\tilde{w}=\tilde{B}^{-1}\Phi \vec{y}/|\B|$. In our toy setup, this can again be done directly in the feature space. In an application with larger feature space, this is still feasible using the matrix-inversion lemma on Eq.~\eqref{eq:batch-Hessian}, instead inverting a dense matrix of size $|\B|\times|\B|$. The Figure also shows the progression of this stochastic estimate (labelled as \emph{avg-inv}, always using $|\B|=256$ since smaller batch-sizes did not work at all). It performs much worse unless the batch-size is increased, which highlights the advantage of the active selection of projection directions for identifying appropriate eigen-vectors. A third option is to use an iterative solver with noise-corrupted observations to approach the optimum. In figure~\ref{fig:ex_linear} a barely visible line labelled CG can be seen which used the method of conjugate gradients with a batch-size of 256. This method required a batch-size $|\B|>10 000$ to show  reasonable convergence on the training objective but would still perform poorly on the test set.

\subsection{Logistic Regression} 
\label{sub:logistic_regression}


Figure~\ref{fig:ex_logistic} shows an analogous experiment on a more realistic, and \emph{non-linear} problem: Classic linear logistic regression on the digits 3 and 5 from the MNIST data-set (i.e.~using linear features $\phi(x)=x$, and $p(y\g x)=\sigma(\phi(x)\Trans w)$). Here we used the model proposed in \citet[][\textsection 3.4]{Rasmussen2005}, which defines a convex, non-linear regularized empirical risk minimization problem that again allows the construction of stochastic gradients, and associated noisy Hessian-vector products. Analogous to Figure~\ref{fig:ex_linear}, Figure~\ref{fig:ex_logistic} shows progress of sgd and pre-conditioned sgd. As before, this problem is actually just small enough to compute an exact solution by Newton optimization (gray baseline in plot). And as before, computation of the pre-conditioner takes up a small fraction of the optimization runtime.

\begin{figure*}[t]
  \centering \scriptsize
  \includegraphics{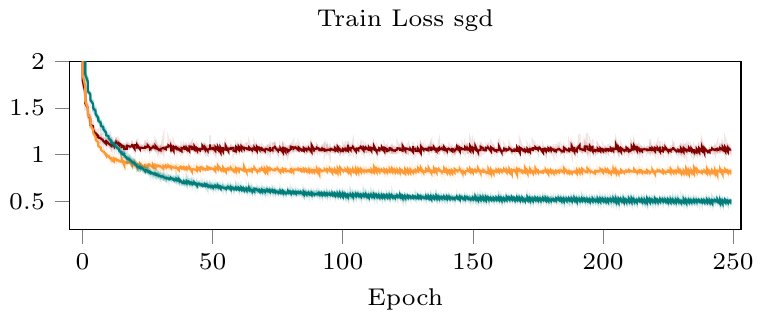}\hfill%
  \includegraphics{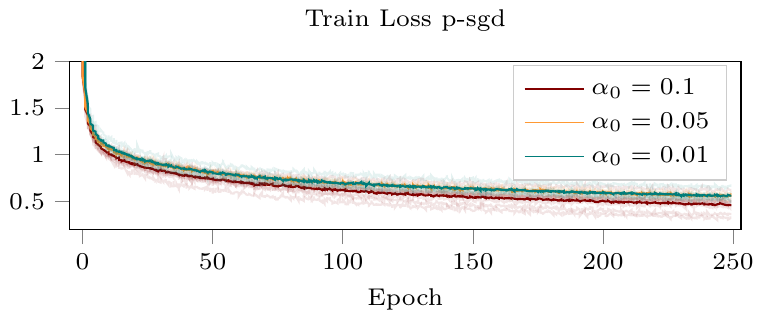}
  \caption{Training loss for sgd (left) and pre-conditioned sgd (right) on the CIFAR-10 data-set for different learning rates and batch-size of 32 over 250 epochs. Both graphs share y-axis and colors to facilitate comparison between the optimizers. The solid lines represent the mean of several individual runs plotted as translucent.}
  \label{fig:dl_train}
\end{figure*}


\subsection{Deep Learning} 
\label{sub:deep_learning}
For a high-dimensional test bed, we used a deep net consisting of convolutional and fully-connected layers (see appendix \ref{app:deep_learning} for details) to classify the CIFAR-10 data-set \citep{krizhevsky2009learning}. The proposed algorithm was implemented in PyTorch \citep{paszke2017automatic}, using the modifications listed in section~\ref{sub:high_dimensional_modification}.

\begin{figure*}[h]
  \centering \scriptsize
  \includegraphics{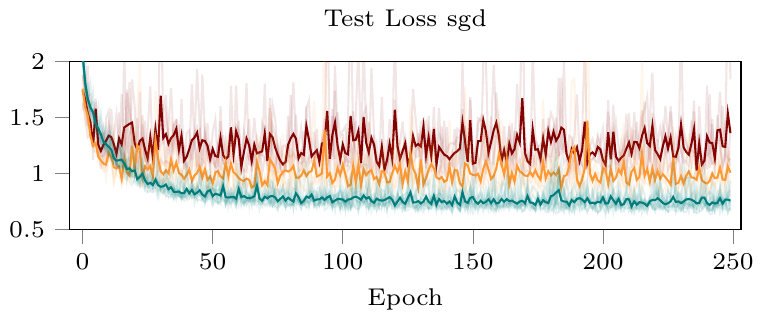}\hfill%
  \includegraphics{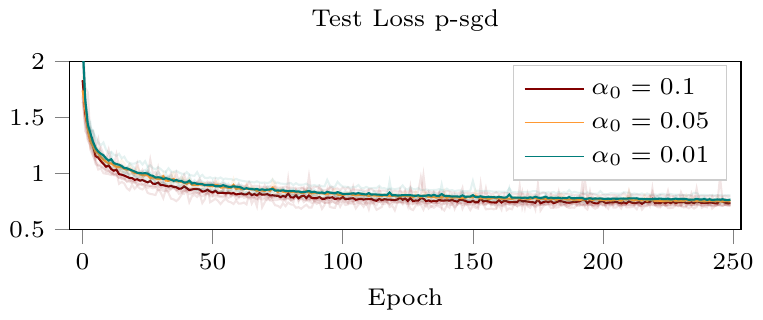}
  \caption{Test loss for sgd (left) and pre-conditioned sgd (right) on the CIFAR-10 data-set for different learning rates and batch-size of 32 over 250 epochs. Both graphs share y-axis and colors to facilitate comparison between the optimizers. These graphs were collected at the same runs as the results in Fig.~\ref{fig:dl_train}.}
  \label{fig:dl_test}
\end{figure*}


To stabilize the algorithm, a predetermined fixed learning-rate was used for the first epoch of the pre-conditioned sgd.
Figure \ref{fig:dl_train} and \ref{fig:dl_test} compare the convergence for the proposed algorithm against sgd for training loss and test loss respectively on CIFAR-10. In both figures we see that the pre-conditioned sgd has similar performance to a well-tuned sgd regardless of the initial learning rate. To keep the cost of computations and storage low we used a rank 2 approximation of the Hessian that was recalculated at the beginning of every epoch. The cost of building the rank 2 approximation was 2--5\% of the total computational cost per epoch.

\begin{figure}
  \centering \scriptsize
  \includegraphics{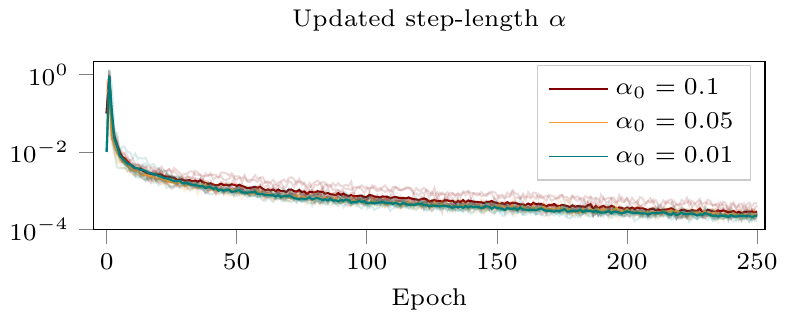}
  \caption{Evolution of the estimated learning rate $\alpha$ over 250 epochs for different initial values.}
  \label{fig:alpha}
\end{figure}

The improved convergence of pre-conditioned sgd over normal sgd is mainly attributed the adaptive step-size, which seems to capture the general scale of the curvature to efficiently make progress. This approach offers a more rigorous way to update the step-length over popular empirical approaches using exponentially decaying learning-rates or division upon plateauing. By studying the scale of the found learning rate, see Fig.~\ref{fig:alpha}, we see that regardless of the initial value, all $\alpha$ follow the same trajectory although spanning values across four orders of magnitude.

\section{CONCLUSION} 
\label{sec:conclusion}

We have proposed an active probabilistic inference algorithm to efficiently construct pre-conditioners in stochastic optimization problems. It consists of three conceptual ingredients: First, a matrix-valued Gaussian inference scheme that can deal with structured Gaussian noise in observed matrix-vector products. Second, an active evaluation scheme aiming to collect informative, non-redundant projections. Third, additional statistical and linear algebra machinery to empirically estimate hyper-parameters and arrive at a low-rank pre-conditioner. The resulting algorithm was shown to significantly improve the behaviour of sgd in imperfectly conditioned problems, even in the case of severe observation noise typical for contemporary machine learning problems. It scales from low- to medium- and high-dimensional problems, where its behaviour qualitatively adapts from full and stable pre-conditioning to low-rank pre-conditioning and, eventually, scalar adaptation of the learning rate of sgd-type optimization.

\subsubsection*{Acknowledgements}
Filip de Roos acknowledges support by the International Max Planck Research School for Intelligent Systems.
Philipp Hennig gratefully acknowledges financial support by the European Research Council through ERC StG Action 757275 / PANAMA.

\newpage
\printbibliography

\appendix
\pagebreak
\begin{center}
\hsize\textwidth
  \linewidth\hsize \toptitlebar {\centering
  {\Large\bfseries Supplementary Material \par}
  \vspace{5pt}
  \large{\textbf{Active Probabilistic Inference on Matrices\\ for Pre-Conditioning in Stochastic Optimization}}}
 \bottomtitlebar \vskip 0.2in plus 1fil minus 0.1in
\end{center}

\begin{minipage}[t]{1.0\textwidth}
  \centering 
  \includegraphics{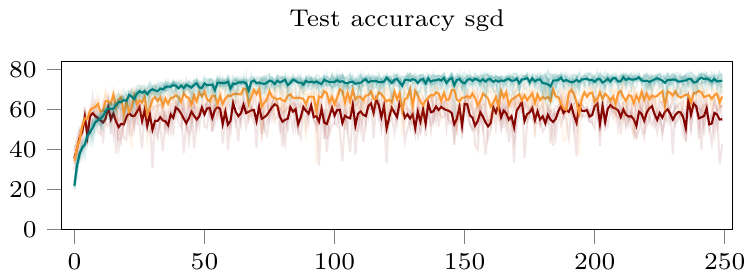}\hfill%
  \includegraphics{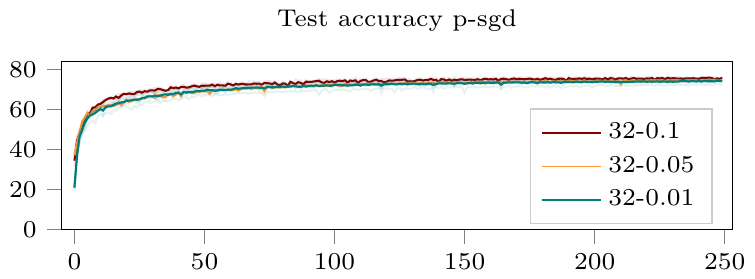}
  \captionof{figure}{Test accuracy for sgd (left) and pre-conditioned sgd (right) on the CIFAR-10 data-set for different learning rates and batch-size of 32 over 250 epochs. Both graphs share y-axis and colors to facilitate comparison between the optimizers. These graphs were collected at the same runs as the results in Fig.~\ref{fig:dl_train}.}\label{fig:dl_test_appendix} 
\end{minipage}


\section{Deep Learning} 
\label{app:deep_learning}
 All deep learning experiments used a neural network consisting of 3 convolutional layers with 64, 96 and 128 output channels of size $5\times 5$, $3\times 3$ and $3\times 3$ followed by 3 fully connected layers of size 512, 256, 10 with cross entropy loss function on the output and $L_2$ regularization with magnitude $0.01$. All layers used the ReLU nonlinearity and the convolutional layers had additional max-pooling.

\end{document}